\documentclass[10pt,technote]{IEEEtran}
\ifCLASSOPTIONcompsoc
  \usepackage[nocompress]{cite}
\else
  \usepackage{cite}
\fi
\ifCLASSINFOpdf
   \usepackage[pdftex]{graphicx}
\else
   \usepackage[dvips]{graphicx}
\fi
\usepackage{amsmath}
\usepackage{amssymb}

\usepackage{epigraph}
\usepackage{algorithmicx}
\usepackage{algorithm}
\usepackage{algpseudocode}
\DeclareMathOperator*{\argmax}{arg\!max}

\usepackage[pagebackref=true,breaklinks=true,letterpaper=true,colorlinks,bookmarks=false]{hyperref}

\begin{document}
%
\title{PsyPhy: A Psychophysics Driven Evaluation Framework for Visual Recognition}
%
%
%
%

\author{Brandon~RichardWebster,~\IEEEmembership{Student~Member,~IEEE,}
        Samuel~E.~Anthony,~\IEEEmembership{Student~Member,~IEEE,}
        and~Walter~J.~Scheirer,~\IEEEmembership{Senior~Member,~IEEE}
\IEEEcompsocitemizethanks{\IEEEcompsocthanksitem B. RichardWebster and W. Scheirer are with the Department of Computer Science and Engineering, University of Notre Dame, Notre Dame, IN, 46556.\protect\\
Corresponding Author's E-mail: brichar1@nd.edu.
\IEEEcompsocthanksitem S. Anthony is with the Department of Psychology, Harvard University, and Perceptive Automata, Inc.}
}

\IEEEtitleabstractindextext{%
\begin{abstract}
By providing substantial amounts of data and standardized evaluation protocols, datasets in computer vision have helped fuel advances across all areas of visual recognition. But even in light of breakthrough results on recent benchmarks, it is still fair to ask if our recognition algorithms are doing as well as we think they are. The vision sciences at large make use of a very different evaluation regime known as Visual Psychophysics to study visual perception. Psychophysics is the quantitative examination of the relationships between controlled stimuli and the behavioral responses they elicit in experimental test subjects. Instead of using summary statistics to gauge performance, psychophysics directs us to construct item-response curves made up of individual stimulus responses to find perceptual thresholds, thus allowing one to identify the exact point at which a subject can no longer reliably recognize the stimulus class. In this article, we introduce a comprehensive evaluation framework for visual recognition models that is underpinned by this methodology. Over millions of procedurally rendered 3D scenes and 2D images, we compare the performance of well-known convolutional neural networks. Our results bring into question recent claims of human-like performance, and provide a path forward for correcting newly surfaced algorithmic deficiencies.
\end{abstract}

}

\maketitle

\IEEEdisplaynontitleabstractindextext

%
\IEEEpeerreviewmaketitle

\vspace{10mm}

\IEEEraisesectionheading{\section{Introduction}\label{sec:introduction}}

%
%
%
%
\epigraph{We often attribute ``understanding" and other cognitive predicates by metaphor and analogy to cars, adding machines, and other artifacts, but nothing is proved by such attributions.}{\textit{John Searle}}

\begin{figure}[ht]
  \centering
    \includegraphics[width=\columnwidth]{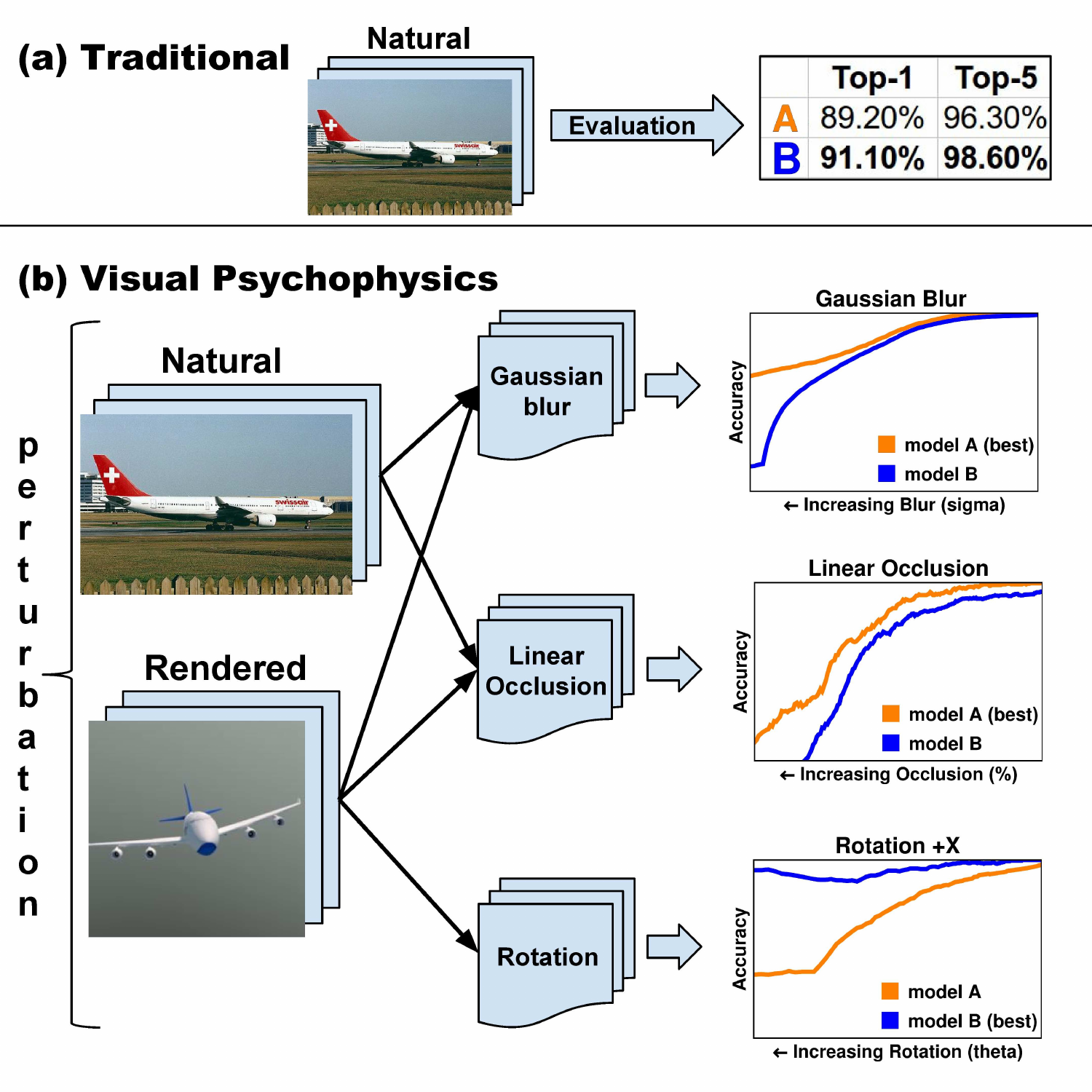}
  \caption{In this article, the concept of applying psychophysics~\cite{lu2013visual,prins2016psychophysics} on a recognition model is introduced. In this figure, A and B are two models being compared. (Top) In traditional dataset-based evaluation, summary statistics are generated over large sets of data, with little consideration given to specific conditions that lead to incorrect recognition instances. (Bottom) Psychophysics, a set of experimental concepts and procedures from psychology and neuroscience, helps us plot the exact relationships between perturbed test images and resulting model behavior to determine the precise conditions under which models fail. Instead of comparing summary statistics, we compare item-response curves representing performance (y-axis) versus the dimension of the image being manipulated (x-axis). }
  \label{fig:teaser}
\end{figure}

Imagine the following scenario: a marvelous black box algorithm has appeared that purportedly solves visual object recognition with human-like ability. As a good scientist, how might you go about falsifying this claim? By all accounts, the algorithm achieves superior performance on established benchmarks in computer vision, but its internal workings are opaque to the external observer. Such a situation is not far fetched --- it should be familiar to any of us studying machine learning for visual recognition. But what many of us in computer vision might not realize is that this setup happens to be the classic Chinese Room~\cite{searle1980minds} problem proposed by the philosopher John Searle. 

In Searle's thought experiment, a person who does not speak Chinese is alone in a locked room and following instructions from a computer program to generate Chinese characters to respond to Chinese messages that are slipped under the door. To the message passer outside of the room, the person inside understands Chinese. However, this is not the case. The person inside the room is simply following instructions to complete the task --- there is no real replication of the competency of knowing the Chinese language. Linking this back to computer vision, the summary statistics of performance from our algorithms look good on benchmark tests --- enough so that we believe them to be close to human performance in some cases. But are these algorithms really solving the general problem of visual object recognition, or are they simply leveraging ``instructions" provided in the form of labeled training data to solve the dataset?

Datasets in computer vision are intended to be controlled testbeds for algorithms, where the task and difficulty can be modulated to facilitate measurable progress in research. A dataset could be made up of images specifically acquired for experimentation, or publicly available images crawled from the web. Under this regime, strong advancements have been demonstrated for a number of problems, most notably for object recognition~\cite{russakovsky2015imagenet}. Deep learning is now a mainstay in computer vision thanks in part to the 2012 ImageNet Challenge~\cite{imagenet2012}, where AlexNet~\cite{NIPS2012_4824} reduced top-5 object classification error to $16.4\%$ from the previously best reported result of $25.8\%$. When algorithms are evaluated on a common footing, it is possible to track meaningful improvements in artificial intelligence like this one. However, increases in error when different datasets are used for training and testing~\cite{torralba2011unbiased, Tommasi2015} make us wonder if this is the \textit{only} way to do  it.  

When it comes to natural intelligence, neuroscientists and psychologists do not evaluate animals or people in the same way that computer vision scientists evaluate algorithms --- and for a very good reason. With a collection of images crawled from the web, there is no straightforward way to determine the exact condition(s) that caused a subject to fail at recognizing a stimulus presented during an experiment. A natural image is the product of the physics at the instant the sensor acquired the scene; its latent parameters are largely unknown. Instead, for behavioral experiments meant to discover perceptual thresholds (\textit{i.e.}, the average point at which subjects start to fail), the vision sciences outside of computer vision use the concepts and procedures from the discipline of \textit{visual psychophysics}.

Psychophysics is the quantitative study of the relationships between controlled stimuli and the behavioral responses they elicit in a subject~\cite{lu2013visual,prins2016psychophysics}. It is a way to probe perceptual processes through the presentation of incremental and, in many cases, extremely fine-grained perturbations of visual stimuli. The properties of each stimulus are varied along one or more physical dimensions, thus controlling the difficulty of the task. The result (Fig.~\ref{fig:teaser}) is an \textit{item-response curve}~\cite{embretson2000item}, where performance (\textit{e.g.}, accuracy) on the y-axis is plotted against the dimension being manipulated (\textit{e.g.}, Gaussian blur) on the x-axis. Each point on the curve reflects an individual stimulus, letting us map performance back to causal conditions in a precise manner. Psychophysics is an indispensable tool to vision science, and has been deployed to discover the minimum
threshold for stimulation of a retinal photoreceptor (a single photon)~\cite{hecht1942energy}, confirm Helmholtz's assertions on color absorption in the retina~\cite{bowmaker1980visual}, and establish criteria to diagnose prosopagnosia~\cite{duchaine2006cambridge} (the inability to recognize a face). As in these discoveries from biological vision, we submit that psychophysics holds much promise for discovering new aspects of the inner workings of machine learning models.

In this article, we introduce a comprehensive evaluation framework for visual recognition that is underpinned by the principles of psychophysics. In this regime, a stimulus can be an object drawn from purely rendered data or natural scene data, and a varying physical parameter can control the amount of transformation in the subsequent set of manipulated images derived from the original stimulus. A key difference from traditional benchmarks in computer vision is that instead of looking at summary statistics (\textit{e.g.}, average accuracy, AUC, precision, recall) to compare algorithm performance, we compare the resulting item-response curves. For complete control of the underlying parameter space, we find that procedural graphics~\cite{tenenbaum2011grow,yildirim2015efficient,Kulkarni_2015_CVPR,Wu_2016} are a useful way to generate stimuli that can be manipulated in any way we desire. Because we have the procedure that rendered each scene, we can find out where a model is failing at the parametric level. As we will see, by using this framework to explore artificial vision systems like psychologists, many interesting new findings can be surfaced about the strengths and limitations of computer vision models.

To summarize, our main contributions are as follows:

\begin{itemize}
    \item A general evaluation framework is developed for performing visual psychophysics on computer vision models. The framework has a strong grounding in well-established work in psychology and neuroscience for behavioral experimentation.
    \item An investigation of procedural graphics for large-scale psychophysics experiments applied to models.
    \item A parallelized implementation of the psychophysics framework that is deployable as a Python package.
    \item A case study consisting of a battery of experiments incorporating millions of procedurally rendered images and 2D images that were perturbed, performed over a set of well-known Convolutional Neural Network (CNN) models~\cite{NIPS2012_4824,jia2014caffe,szegedy2015going,DBLP:journals/corr/SimonyanZ14a}.
    
\end{itemize}

\section{Related Work}

\textbf{Methods of Evaluation from the Vision Sciences.} 
With respect to work in computer vision directly using psychophysics, most is related to establishing human baselines for comparison to algorithmic approaches.
Riesenhuber and Poggio~\cite{riesenhuber2000individual} described a series of psychophysical comparisons between humans and the HMAX~\cite{riesenhuber1999hierarchical} model of visual cortex using a limited set of stimuli rendered by computer graphics. 
Similarly, Eberhardt et al.~\cite{eberhardt2016deep} designed an experiment to measure human accuracy and reaction time during visual categorization tasks with natural images, which were then compared to different layers of CNN models~\cite{NIPS2012_4824,DBLP:journals/corr/SimonyanZ14a}. Geirhos et al. undertook a similar study for image degradations~\cite{DBLP:journals/corr/GeirhosJSRBW17}. With respect to low-level features, Gerhard et al.~\cite{gerhard2013sensitive} introduced a new psychophysical paradigm for comparing human and model sensitivity to local image regularities. 

Psychophysics can also be used for more than just performance evaluation. Scheirer et al.~\cite{Scheirer_2014_TPAMIa} introduced the notion of ``perceptual annotation" for machine learning, whereby psychophysical measurements are used as weights in a loss function to give a training regime some \textit{a priori} notion of sample difficulty. Using accuracy and reaction time measured via the online psychophysics testing platform TestMyBrain.org~\cite{germine2012web}, perceptual annotation was shown to enhance face detection performance. Along these lines, Vondrick et al.~\cite{NIPS2015_5781} devised a method inspired by psychophysics to estimate useful biases for recognition in computer vision feature spaces.     

Outside of work specifically invoking psychophysics, one can find other related methods from psychology and neuroscience for behavioral-style model testing. 2D natural images are the most common type of data in computer vision, and form a good basis for algorithmic evaluation in this mode.
O'Toole et al.~\cite{o2007face,o2012comparing} and Philips et al.~\cite{phillips2014comparison} have designed controlled datasets of natural images to compare human face recognition performance against algorithms. With the focus on algorithmic consistency with human behavior, there is no explicit model vs. model comparison in these methods.

 More control in experimentation can be achieved through the use of rendered 3D scenes. Cadiue et al.~\cite{cadieu2013neural}, Yamins et al.~\cite{yamins2014performance} and Hong et al.~\cite{hong2016explicit} all make use of rendered images with parametrized variation to compare the representations of models with those found in the primate brain. Pramod and Arun~\cite{Pramod_2016_CVPR} describe a set of perceived dissimilarity measurements from humans that is used to study the systematic differences between human perception and a large number of handcrafted and learned feature representations. Because of a need for very fine-grained  control of object parts and other latent parameters of scenes, procedural graphics were introduced by Tenenbaum et al.~\cite{tenenbaum2011grow} for the study of one-shot learning using probabilistic generative models. The use of procedural graphics for generative models was further developed by Yildirim et al.~\cite{yildirim2015efficient}, Kulkarni et al.~\cite{Kulkarni_2015_CVPR}, and Wu et al.~\cite{Wu_2016}. These studies do not vary the conditions of the stimuli using the procedures of psychophysics, nor do they use large-scale renderings on the order of millions of scenes. 

\textbf{Other Manipulations of Stimuli in Visual Recognition Evaluations.} 
Work coming directly out of computer vision also addresses stimulus generation for the purpose of isolating model weaknesses. Hoiem et al.~\cite{hoiem2012diagnosing} suggest systematically varying  occlusion, size, aspect ratio,
visibility of parts, viewpoint, localization error, and background to identify errors in object detectors. Wilber et al.~\cite{wilber2016can} systematically apply noise, blur, occlusion, compression, textures and warping effects over 2D scenes to assess face detection performance.
Finally, a whole host of approaches can be found to manipulate the inputs to CNNs in order to highlight unexpected classification errors. These include the noise patterns introduced by Szegedy et al.~\cite{szegedy2013intriguing} that are imperceptible to humans, and the fooling images produced via evolutionary algorithms that were explored by Nguyen et al.~\cite{nguyen2015deep} and Bendale and Boult~\cite{Bendale_2016_CVPR}. The level of control in the  evaluation procedures varies between these approaches, but a common starting point based on model preference for each class is missing (\textit{i.e.}, which object configuration produces the highest score?). We suggest in this article that the use of computer graphics helps us address this. 

\section{Method: The PsyPhy Framework}
\label{sec:method}
Our procedure for performing psychophysics on a model largely follows established procedures found in psychology, with a few key adaptations to accommodate artificial perception.  For the purposes of this article, our focus is on two performance-based forced-choice tasks that yield an interpretable item-response curve. For descriptions of other procedures in psychophysics, see~\cite{lu2013visual,prins2016psychophysics}. First, let us consider the  \textit{two-alternative forced choice (2AFC) match-to-sample} task that is common in psychological testing. 

In the 2AFC match-to-sample procedure, an observer is shown a ``sample" stimulus, followed by two ``alternate" stimuli where one is a positive (\textit{i.e.}, matching) stimulus and the other is a negative (\textit{i.e.}, non-matching) stimulus. The observer is then asked to choose from the alternate stimuli the stimulus that best matches the sample --- the match criterion may or may not be provided to the observer. The observer repeats the task at different perturbed stimulus levels in either an adaptive pattern, which is like gradient descent for humans, or via the method of constants, which is a predetermined set of perturbed stimulus levels. Regardless of method, each task has two presented alternate stimuli ($N=2$) and thus two-alternative forced-choices ($M=2$). 
Analysis of the experiment would utilize the mean or median accuracy humans achieved at each stimulus level and mean or median human response time, if recorded. Models can be tested in precisely the same way when the input images are arranged as described above and accuracy is used as the performance measure.

Second, we can consider a mapping of a more difficult classification procedure in machine learning to a more general version of the 2AFC match-to-sample procedure. We call this mapped classification \textit{MAFC  match-to-sample}. In MAFC, the probe image in classification is equivalent to the sample stimulus.
In classification, we rarely have only two classes for a model to choose from. Thus the value of $M$  becomes the number of labeled training classes (\textit{e.g.}, ImageNet 2012 has $1$K learned classes, making $M=1$K). Likewise, $N$ --- the number of presented alternate stimuli --- changes to the number of images used in training, as this is the set of images the model is implicitly matching to (\textit{e.g.}, for  ImageNet 2012, $N = {\sim}1.2$M training images).

When testing a model with any psychophysics procedure, we need a special process for the selection of the stimuli's default state, that is, where there is no perturbation. Blanz et al.~\cite{blanz1999object} show that humans have a natural inclination towards recognizing a certain view of an object, called a canonical view. We assume in human trials that an object configuration close to a canonical view will be chosen, maximizing the probability that all observers will have no problems performing at least some part of a match-to-sample task. However, this is not as simple for any task involving a model because we do not necessarily know if it follows a similar canonical view. But we can say that a model's \textit{preferred view} is a view that produces the strongest positive response, as 
determined by a decision score. Note that there can be more than one preferred view (hence our use of the term \textit{preferred}), because ties are often observed for the strongest response, especially in quantized decision score spaces. 
Choosing a preferred view is crucial to guaranteeing that when the stimulus is perturbed, the model's response will already be at its maximum for that class. Any perturbation will cause a decline (or possibly no change) in the strength of the model's response, not an increase.

\begin{algorithm}[t]
    \begin{algorithmic}[1]
        \Require{$f$, a single pre-trained network model}
        \Require{$i$, an input image}
        \Require{$c$, the expected class}
        \State{$V$ = $f(i)$ \Comment{the Softmax vector}}
        \State{$c^* = \argmax_{j \in [0,|V|)} V_{j}$ \Comment{find class label}}
        \State{$\varsigma = V_{c^*}$}
        \If {$c \neq c^*$} \Comment{incorrect class, negate response}
            \State{$\varsigma = -1 * \varsigma$}
        \EndIf \\
        \Return{$\varsigma$, the decision score}
    \end{algorithmic}
    \caption{$D_{f}^{m} (i,c)$, the top-1 binary decision of the Softmax layer of a CNN. Used for both preferred view calculation and MAFC.}
        \label{alg:dec2}
\end{algorithm}

\textbf{PsyPhy Framework for 2AFC and MAFC.} Inspired by software frameworks for subject testing like PsychoPy~\cite{Peirce20078}, we have implemented the 2AFC and MAFC procedures described above using a Python framework for model testing called PsyPhy. Here we describe the details of each component of the framework. The basic steps of (1) stimuli selection, (2) preferred view selection, (3) perturbation, and (4) item-response curve generation apply to any psychophysics procedure, and the specific 2AFC and MAFC procedures may be viewed as pluggable modules within the framework. PsyPhy is very flexible with respect to tasks it can support.  

The first step is to select the initial set of stimuli for each class. For 2D natural images, this is any set of chosen images $I_{2D}$ for a class $c$. For a rendered scene, a set of image specifications $I_{3D}$ is provided to a rendering function $R(c,v)$ (implemented in this work using Mitsuba~\cite{Mitsuba}) to render a single object centered in an image. The view $v\in I_{3D}$ is the parameter set $\{x,y,z,\psi\}$, where the coordinates $x$, $y$, and $z$ are real numbers in the range $(-180.0,180.0]$ and $\psi$, representing scale, is a real number in the range $(0.0, 25.0]$.

\begin{algorithm}[t]
    \begin{algorithmic}[1]
        \Require{$D_f^2$, decision function for 2AFC}
        \Require{$V$, a vector of preferred views for a set of classes}
        \Require{$s$, the stimulus level}
        \State{$h(v) := $\textbf{random $v'$}$|v'\in [V \setminus \{v\}] $ } \Comment{pick negative}
        \State{$\beta = \sum_{v \in V} \max (0,\lceil D_f^2(T(s,v),v, h(v)) \rceil)$}
        \State{$a = \frac{\beta}{|V|}$} \\
        \Return{$\{s, a\}$, an $x,y$ coordinate pair (stimulus level, accuracy over trials) for one item-response point}
    \end{algorithmic}
    \caption{$\phi_{T}^2 (D_f^2,V,s)$: an item-response point generation function supporting 2AFC tasks for any image transformation function $T(s,v)$}
        \label{alg:per1}
\end{algorithm}

The second step is to find an individual model's preferred view for each class.
For natural 2D images, the preferred view function in Eq.~\ref{equ:pnat} is used. The second preferred view function, Eq.~\ref{equ:pren}, uses $R$ to create rendered images for classification. In Eq.~\ref{equ:pren}, the search space is almost infinite, thus it does not find the absolute global maximum, but rather an approximation. 
\begin{equation} \label{equ:pnat}
{\cal P}_{2D} (I_{2D}, c) := \argmax_{i\in I_{2D}} D^m_f(i, c)
\end{equation}
\begin{equation} \label{equ:pren}
{\cal P}_{3D} (I_{3D}, c) := \argmax_{v\in I_{3D}} D^m_f(R(c, v), c)
\end{equation}
A decision function for classification $D^m_f(i,c)$ (Alg.~\ref{alg:dec2}) normalizes the score output of a model $f$ to a value in the range $[-1.0, 1.0]$, which gives both a decision and a confidence associated with that decision. A value in the range $[-1.0,0]$ is an incorrect decision and $(0,1.0]$ is a correct decision. The parameter $i$ is the input stimulus and $c$ is the expected class.

A natural 2D preferred view (Eq.~\ref{equ:pnat}) is a single selected image $i \in I_{2D}$, where $D_f^m$ has the strongest positive response. A 3D preferred view (Eq.~\ref{equ:pren}) is a single selected set ${v = \{x_v,y_v,z_v,{\psi}_v\}} \in I_{3D}$, where $D_f^m$ has the strongest positive response. The major difference between Eq.~\ref{equ:pnat} and Eq.~\ref{equ:pren} is the use of  $R$ in Eq.~\ref{equ:pren} to render the image prior to measuring the response from $D_f^m$. Invoking Eq.~\ref{equ:pnat} or Eq.~\ref{equ:pren} for each class will create a vector of preferred views $V$.

After preferred views have been selected for all classes, whether natural or rendered, the next step is to apply perturbations to them. In this procedure, a set of preferred views is perturbed at a specific stimulus level (\textit{i.e.}, the amount of perturbation) using a function $T(s,v)$, where $T$ could be any image transformation function (\textit{e.g.}, Gaussian blur, rotation). The parameter $v$ is one  preferred view --- either in 2D image format or $\{x,y,z,\psi\}$ for rendered stimuli --- and $s$ is the stimulus level. The function $\phi_{T}(D,V,s)$ perturbs the set of preferred views given in $V$ and then makes a decision on each image using a decision function $D$. The specific implementation $\phi^2_{T}(D_f^2,V,s)$ for 2AFC is described in Alg.~\ref{alg:per1}, and $\phi_{T}^m(D_f^m,V,s)$ for MAFC is described in Alg.~\ref{alg:per2}. Procedure specific decision functions are required, with $D_f^2$ (Alg.~\ref{alg:dec1}) used for 2AFC and $D_f^m$ (Alg.~\ref{alg:dec2}) used for MAFC.  Each individual image evaluation is a trial. The value returned by $\phi_{T}$ represents one point on an item-response curve, which is the computed accuracy over all trials (one trial per class).




\begin{algorithm}[t]
    \begin{algorithmic}[1]
        \Require{$D_f^m$, decision function for MAFC}
        \Require{$V$, a vector of preferred views for a set of classes}
        \Require{$s$, the stimulus level}
        \State{$\beta = \sum_{v \in V} \max (0,\lceil D_f^m(T(s,v),c(v)) \rceil)$}
        \State{$a = \frac{\beta}{|V|}$} \\
        \Return{$\{s, a\}$, an $x,y$ coordinate pair (stimulus level, accuracy over trials) for one item-response point}
    \end{algorithmic}
    \caption{$\phi_{T}^m (D_f^m,V,s)$: an item-response point generation function supporting MAFC tasks for any image transformation function $T(s,v)$}
        \label{alg:per2}
\end{algorithm}

An item-response curve is the set of $x,y$ coordinates that represent the model behavior for a set of stimuli. Each $x,y$ value represents a perturbation level and accuracy of the model's performance. Note that traditional psychophysics experiments with live test subjects often apply a  psychometric function to interpolate between the points to generate the curve. To approximate a psychometric function for better interpretability, we applied rectangular smoothing (\textit{i.e.}, unweighted sliding-average smoothing) with a window size of $15$ while padding the curve with repeated edge values. 

The final step generates item-response curves using the function ${\cal C}_T (\phi,D,V,n,b_l,b_u)$. The procedure is simple, and only requires a repeated execution of $\phi_{T}$ for each stimulus level. Its steps are shown in Alg.~\ref{alg:all}. The procedure will create a set of stimulus levels starting with a lower bound, $b_l$, and ending with an upper bound $b_u$. $b_l$ is the closest stimulus level to the preferred view and $b_u$ is the stimulus level that is farthest away. The parameter $n$ is the number of stimulus levels to use. Typically in visual pyschophysics, log-spaced stepping is used for finer-grained evaluation near the canonical view; the same strategy is used for preferred view.

\section{Experiments}
The first goal of our experiments was to demonstrate PsyPhy as a large-scale psychophysics evaluation framework. To do this, we processed millions of procedurally rendered 3D scenes and 2D images that were perturbed.  The second goal was to demonstrate the utility of procedural graphics  for large-scale psychophysics experiments. Thus we broke our data up into two sets: natural scenes and rendered scenes. Our final goal was to evaluate the strengths and weaknesses of well-known CNN models. To do this, we looked at model behavior for 2AFC and MAFC tasks, the behavior of dropout at test time~\cite{pmlr-v48-gal16} under perturbations, and comparisons to human behavior. In all of our experiments, we chose to use five  convolutional neural network models that were pre-trained on ImageNet 2012~\cite{imagenet2012}: AlexNet~\cite{NIPS2012_4824}, CaffeNet~\cite{jia2014caffe}, GoogleNet~\cite{szegedy2015going}, VGG-16, and VGG-19~\cite{DBLP:journals/corr/SimonyanZ14a}. The complete set of plots and more details on the methods can be found in the supplemental material for this article\footnote{Supplemental material is accessible at \\ \url{http://bjrichardwebster.com/papers/psyphy/supp}}.


\begin{algorithm}[t]
    \begin{algorithmic}[1]
        \Require{$f$, a single pre-trained network model}
        \Require{$i$, an input image}
        \Require{$p$, the expected positive image}
        \Require{$q$, the expected negative image}
        \State{$W_{i}$ = $f(i)$ \Comment{gather activations from final feature layer}}
        \State{$W_{p}$ = $f(p)$}
        \State{$W_{q}$ = $f(q)$}
        \State{$\varsigma_{p} = r(W_{i}, W_{p})$ \Comment{Pearson's Correlation}}
        \State{$\varsigma_{q} = r(W_{i}, W_{q})$ }
        \If {$\varsigma_{p} > \varsigma_{q}$} \Comment{if incorrect selection, negate response}
            \State{$\varsigma = \varsigma_{p}$}
        \Else 
            \State{$\varsigma = -1 * \varsigma_{q}$}
        \EndIf \\
        \Return{$\varsigma$, the decision score}
    \end{algorithmic}
    \caption{$D_{f}^2 (i,p,q)$, best match decision of the final feature layer of a CNN. Used for 2AFC.}
        \label{alg:dec1}
\end{algorithm}

\begin{algorithm}[t]
    \begin{algorithmic}[1]
        \Require{$\phi$, an item-response point generator}
        \Require{$f$, an input model}
        \Require{$V$, a vector of preferred views}
        \Require{$n$, the number of stimulus levels}
        \Require{$b_l$ and $b_u$, the lower and upper bound values of the stimulus levels}
        \State{\textbf{Let} $S$ be $n$ log-spaced stimulus levels from $b_l$ to $b_u$}
        \State{$I = \bigcup\limits_{s \in S} \{\phi_{T} (D, V, s)\}$} \\
        \Return{$I$, the item-response curve}
    \end{algorithmic}
    \caption{${\cal C}_T (\phi,D,V,n,b_l,b_u)$: an item-response curve generation function for any type of decision function}
        \label{alg:all}
\end{algorithm}

\textbf{Data Generation.} For the natural scene experiments, we perturbed images from the ImageNet 2012~\cite{imagenet2012} training dataset, which consists of ${\sim}1.2$M million images and $1$K classes. Using the training set instead of the testing set gives each model an intentional bias towards ``expert" performance. The following transformations were applied: Gaussian blur, linear occlusion, salt \& pepper noise, brightness, contrast, and sharpness. For each condition, we created $200$ perturbed images starting with the preferred view and log-spaced stepped towards increasing difficulty. The result was $201$ images per class per network, or $201$K images per network, or ${\sim}1$M images per condition. In total, ${\sim}9$M images were evaluated. 




For the experiments with rendered images, we selected $40$ 3D objects from the Blend Swap~\cite{blendswap} library that corresponded to classes in ImageNet (see supp. material for a list of the classes). For each of the 3D objects, we randomly rendered $100$K uniformly distributed $x,y,z$ rotations and scales, resulting in $4$M images. After each preferred view was selected from that set, the following transformations were applied by our graphics engine: rotations in the $x,y,z$ dimensions, and scale. All were applied in the positive and negative direction. In addition,  all of the transformations from the 2D natural image experiment were repeated using the rendered preferred views. For each of the 3D transformations, we rendered $200$ images starting with the preferred view and log-spaced stepped towards increasing difficulty. The result was $201$ images per class per network, or ${\sim}8$K images per network, or ${\sim}40$K images per transformation. The additional 2D transformations resulted in a total of ${\sim}362$K images, which brought the rendered image total to ${\sim}683$K evaluated images.





\begin{figure*}[t]
 \centering
    \includegraphics[width=\textwidth]{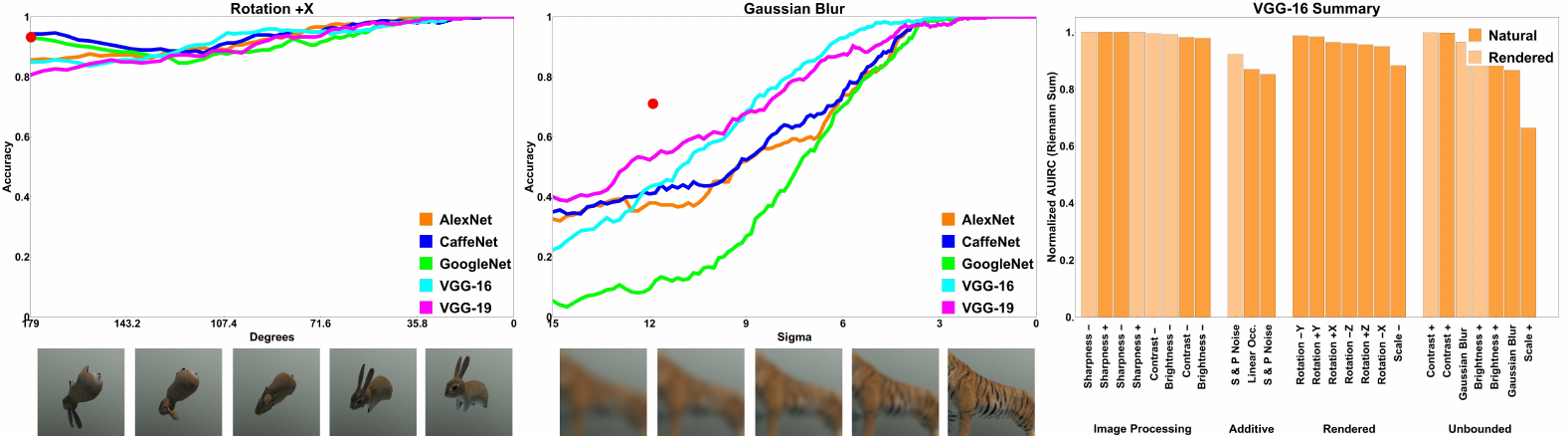}
    \caption{(Left and Center) a selection of item-response curves for the 2AFC task. These rendered scene experiments reflect the accuracy across $40$ classes. Each experiment used five well-known CNNs~\cite{NIPS2012_4824,jia2014caffe,szegedy2015going,DBLP:journals/corr/SimonyanZ14a}. A perfect curve would be a flat line at the top of the plot. The images at the bottom of each curve show how the perturbations increase from right to left, starting with no perturbation (\textit{i.e.}, the original image) for all conditions.
    The red dot indicates mean human performance for a selected stimulus level. (Right) a summary plot of all VGG-16 2AFC item-response curves using AUIRC (Riemann sum) as a summary statistic, normalized by the total area above and below the curve. These plots (as well as the next sets in Figs.~\ref{fig:mafc} \&~\ref{fig:dropout}) are best viewed in color.}
 \label{fig:2afc}
 \vspace{-4mm}
\end{figure*}

\textbf{2AFC Experiments.} The motivation for the 2AFC experiments is twofold: (1) to test decision making at a fundamental level via activation matching (\textit{i.e.}, to not just look at class labels), and (2) to test a precise implementation of a well-known match-to-sample task. Given that this setting is just two instances of pair-matching, we had initially expected models to perform relatively well under a variety of perturbations.  The experiments included the use of both natural scenes and rendered scenes as stimuli.


Model behavior was only very stable for the 3D rotation (Fig.~\ref{fig:2afc} left and Supp. Fig. 1), and contrast and sharpness (Supp. Figs. 3 \& 4) transformations. The rest of the transformations induced more erratic behavior, with accuracy declining below $80\%$. For example, Gaussian blur (Fig.~\ref{fig:2afc} center) was very detrimental to model accuracy, even in this limited matching setting. This tells us something about the receptive fields of the convolutional layers of the networks: they are not large enough to tolerate even modest levels of blur.  Also interesting were results for the 3D-specific scale perturbation that let us isolate specific failure modes. For example, when scale decreases to 20\% of the original size, the object structure is still clearly visible, but accuracy drops to ${\sim}60\%$ or lower for all networks (Supp. Fig. 1). This is an observation that could not have been made by looking at summary statistics. 

What about the differences in behavior across networks? Are they significant? When examining the item-response curves with the the 95\% confidence interval plotted (Supp. Figs. 2, 5, 6 \& 8) all network behavior consistently demonstrates the same trends for each transformation type across perturbations. While it is commonly believed that architectural innovation is advancing deep learning, this finding indicates that model behavior is more influenced by the training data, which was the same for all models in these experiments. For the VGG networks, this suggests that additional layers --- beyond a certain point --- do not imply better performance under degrading conditions. Likewise, switching the order of the pooling and normalization layers in CaffeNet and AlexNet~\cite{caffenet} does not imply better performance under degrading conditions.

\begin{figure*}[t]
 \centering
    \includegraphics[width=\textwidth]{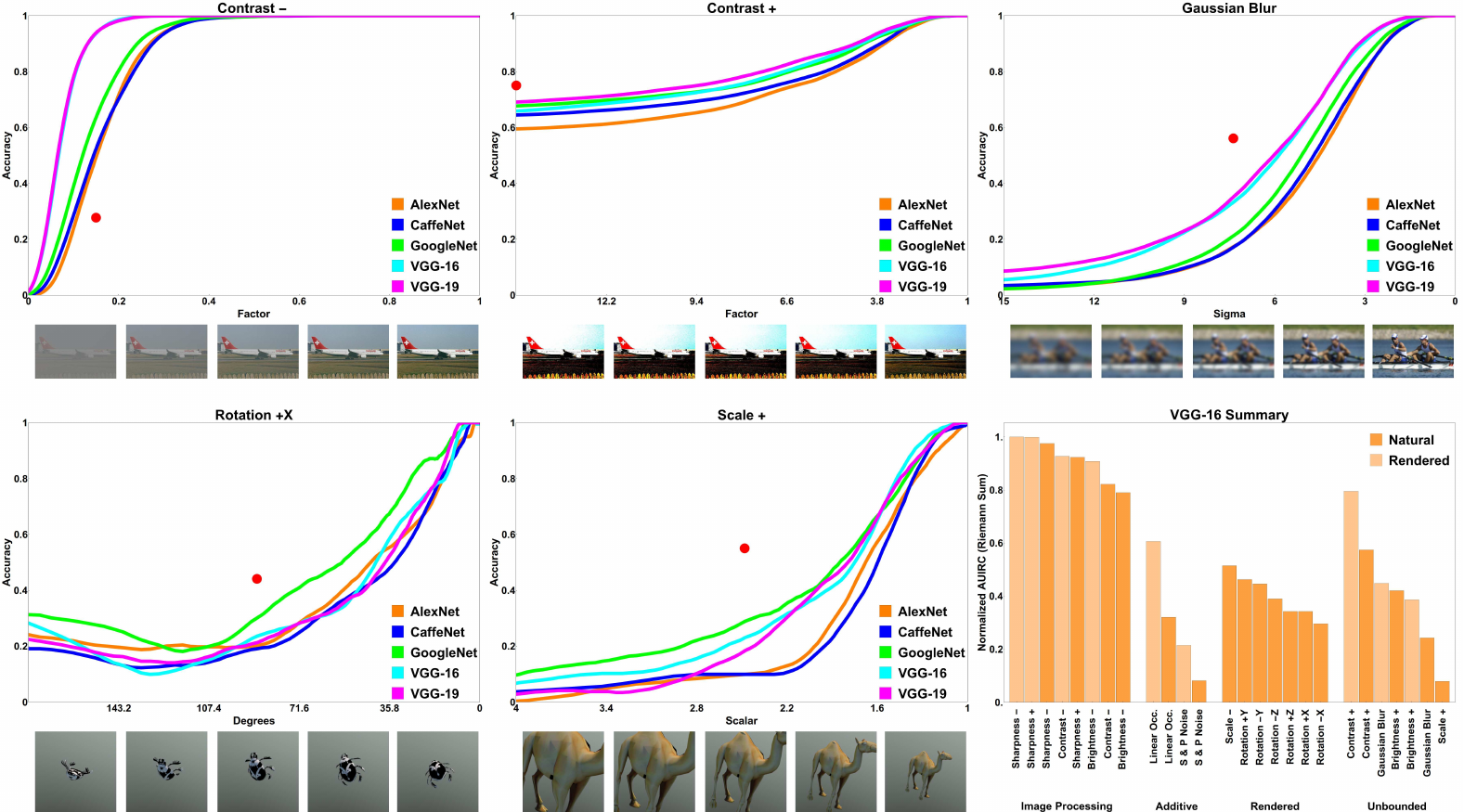}
 \caption{A selection of item-response curves for the MAFC task. (Top) natural scenes. (Bottom-Left and Center) rendered scenes. The top-left can be directly compared to the top-center as well as the bottom-left to its corresponding plot in Fig.~\ref{fig:2afc}. (Bottom-Right) MAFC summary plot for VGG-16.}
 \vspace{-4mm}
 \label{fig:mafc}
\end{figure*}

\textbf{MAFC Experiments.} The motivation for the MAFC experiments is to evaluate a task that is more closely aligned to the multi-class classification task the models were originally trained for. Given that there are $1,000$ choices in this setting instead of just two, we expected models to perform much worse under the transformations. And this is exactly what we observed in the results (Fig.~\ref{fig:mafc} and Supp. Figs. 9, 11, 12 \& 15). For instance, compare the plot for positive rotation in the x-axis (Fig.~\ref{fig:mafc} bottom-left) to the corresponding plot in Fig.~\ref{fig:2afc}. For this transformation type, the networks that only experienced moderate dips in performance for the most extreme perturbations in the 2AFC case fall to under ${\sim}20\%$ accuracy at points. A caveat to the MAFC decision function is that because it is patterned after the classification task in computer vision, it only uses class labels to make its decisions. Thus it leaves out the layer-specific activation information that was used in the 2AFC case. This highlights an important trade-off that can occur when designing decision functions for psychophysics evaluations: task fidelity versus task difficulty. 

Curiously, there are large asymmetries for some of the transformations with increasing and decreasing perturbation levels. See the plots for brightness, contrast, and sharpness (Fig.~\ref{fig:mafc} top-left and top-center, and Supp. Figs. 11 \& 12). Contrast is a particularly intriguing case. As a transformation, contrast is a non-linear single pixel-level operation applied globally to an image. In the positive direction, contrast is increased, and the performance of each network degrades rapidly (Fig.~\ref{fig:mafc} top-center). In the negative direction, contrast is decreased, but the performance of each network remains relatively stable until the objects have very low contrast (Fig.~\ref{fig:mafc} top-left). This suggests a contrast sensitivity problem under the MAFC decision function that is the opposite of what human patients with contrast-related vision deficits struggle with. There is a positive aspect to this finding --- while diminished contrast sensitivity may induce night-blindness in a human driver, CNN-based autonomous driving systems can be expected to operate more effectively in the dark.   

\textbf{Cross-Perturbation Comparison.} To facilitate comparison across perturbations, we generated one summary plot for each set of 2AFC (Fig.~\ref{fig:2afc} right) and MAFC (Fig.~\ref{fig:mafc} bottom-right) experiments. Each plot is generated using an area under the item-response curve (AUIRC) summary statistic, calculated with a midpoint Riemann sum and then normalized to unit space. This is similar in spirit to area under the curve in an ROC setting. A bar representing perfect performance has $y=1.0$. A benefit of using AUIRC allows comparisons across perturbations without making assumptions about the underlying shape of the item-response curve. While model performance can effectively be compared using AUIRC, caution should be taken when comparing unbounded parameters (\textit{e.g.}, $\sigma$ for Gaussian blur) as such a comparison is dependent on the selected bound the experimenter has chosen.

\textbf{Dropout Experiments.} The experiments we have looked at thus far assume deterministic outputs. What about settings with stochastic outputs that support uncertainty in decision making? Gal and Ghahramani~\cite{pmlr-v48-gal16} introduced dropout \textit{at testing time} to implement Bayesian inference in neural networks. What sort of variability does this lead to under various transformations and perturbation levels, and what does this tell us about the certainty of the experiments above? The setup for these experiments is identical to the setup for the MAFC experiments (including preferred views) except that during evaluation we applied dropout at test time to the Caffe version of AlexNet (which was also trained with dropout). Deploying the pre-trained model for each test, we dropped out $50\%$ (because the model is large) of the neurons in layers \textit{fe6} and \textit{fe7} by uniformly randomly setting their activations to zero. This is repeated with $5$ different random seeds for each transformation except salt \& pepper noise and linear occlusion, which were not performed due to randomness in their underlying perturbation functions. 

\begin{figure}[t]
 \centering
    \includegraphics[width=0.85\columnwidth]{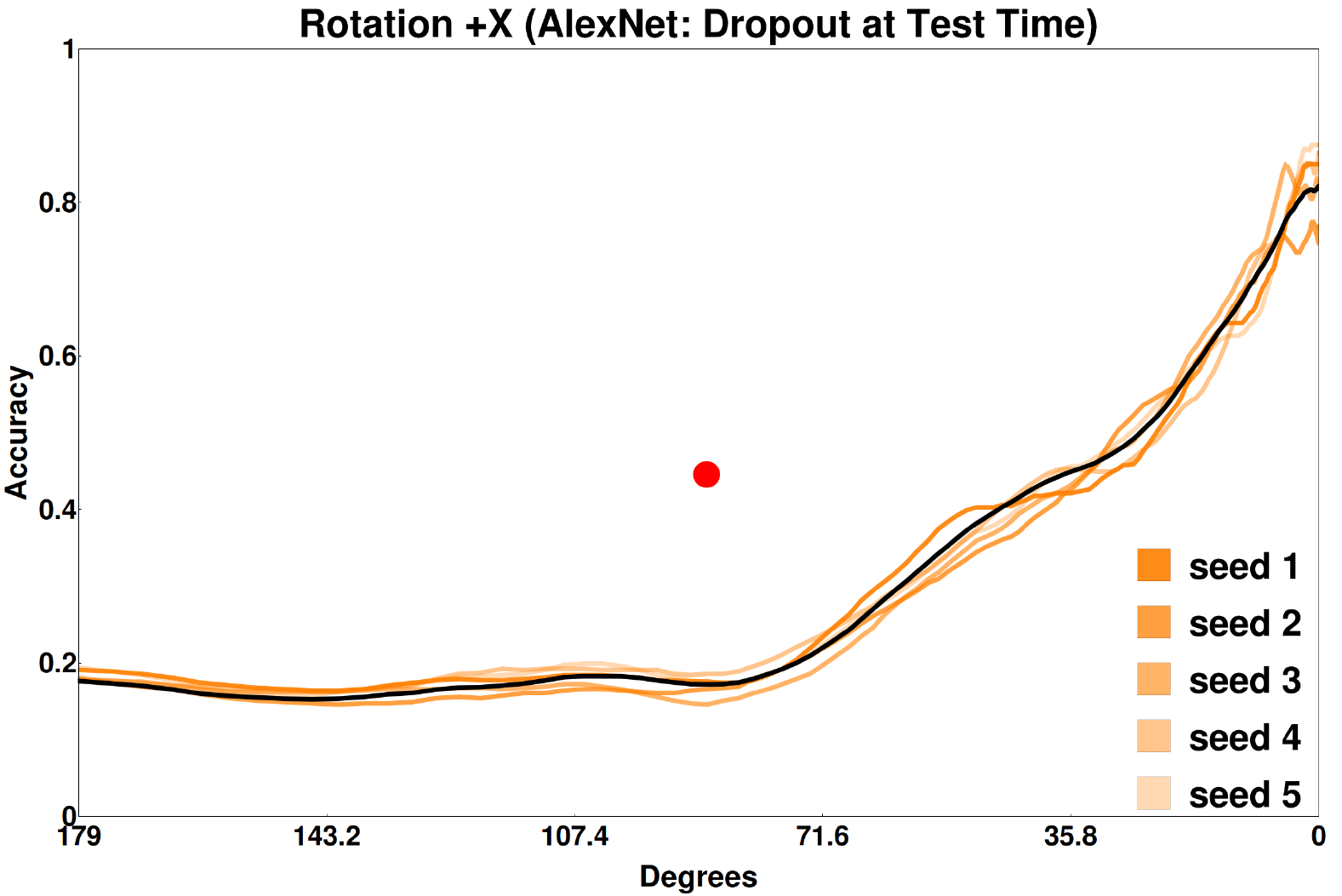}
 \caption{Item-response curves for five different runs of an AlexNet model with dropout applied at test time~\cite{pmlr-v48-gal16} for a 3D rotation transformation. The black line indicates the mean of the five AlexNet curves. The maximum difference  between points on any two curves in this plot is 12.2\%.}
 \vspace{-4mm}
 \label{fig:dropout}
\end{figure}

As anticipated, some variability in the base model performance was introduced (Fig.~\ref{fig:dropout} and Supp. Figs. 17-20). But importantly, most runs still demonstrated a large measure of consistency (\textit{e.g.}, Fig.~\ref{fig:dropout}) across the ranges of perturbations, indicating higher degrees of model certainty. This is a good stability property --- when a model fails, it does so in an expected way across different dropout configurations, lending credibility to the initial characterizations of the behavior in the earlier experiments. More variability was observed for the rendered objects versus the natural scenes. This can be attributed to the use of the 3D objects that were outside of the training set for all of the models. The maximum difference observed between points from two runs for any transformation was $16.5$\% for sharpness applied to 3D objects (Supp. Fig. 19). In over half the cases, the maximum difference was over $10$\%. 


\textbf{Human Comparisons.} Using psychophysics, model performance can be directly compared to human performance. To obtain human data points (red dots) for Figs.~\ref{fig:2afc}-\ref{fig:dropout}, we conducted a study with $24$ participants ($21$ for contrast).  Each participant performed the 2AFC task as described above, but to mitigate fatigue, only $5$ trials with a fixed psychophysical parameter setting were given for each of the transformations from Fig.~\ref{fig:2afc}. The participants also performed the MAFC task, but were limited to $3$ choices instead of the full $1000$ classes to make the task tractable. For those experiments, participants performed in $5$ trials with a fixed psychophysical parameter setting for each transformation in Fig.~\ref{fig:mafc}. The original images for each trial were chosen randomly from the VGG-16 preferred views such that each class was only used one time for each participant in order to prevent participants from learning while performing the task. For all trials on both tasks, the sample images were presented for $50$ms and subjects had unlimited time to answer. 

Even without generating a full psychometric curve for the human subjects, it was apparent that only two out of eleven experiments showed any relative consistency between human and model performance (Fig.~\ref{fig:2afc} left and brightness increasing in Supp. Fig. 1). While human performance was superior to model performance in most cases, there were two cases where humans were worse than the models: decreasing contrast (Fig.~\ref{fig:mafc} left; for analysis, see MAFC experiments) and increasing brightness (Supp. Fig. 3). Brightness adjustment in image processing is a constant multiplicative change to all the pixel values, which preserves edges and allows the networks to recognize the geometry of an object almost until saturation is reached. Humans were also good at this task, but were still ${\sim}9\%$ worse than VGG-19 for the perturbation level analyzed.

\section{Discussion}
In visual psychophysics, we have a convenient and practical alternative to traditional dataset evaluation. However, the use of psychophysics testing and datasets are not mutually exclusive.  One needs datasets to form a training basis for any data-driven model. Moreover, there is major utility to having a large amount of such data --- this is essential for making machine learning capture enough intraclass variance to generalize well to unseen class instances. Data augmentation~\cite{NIPS2012_4824,chatfield2014return} is an obvious strategy for leveraging the rendered images that were problematic for a model during psychophysics testing to expand the scope of the training set. However, this has diminishing returns as datasets grow to sizes that exceed available memory (or even disk space) during training. Using more limited training data and reinforcement learning that optimizes over item-response curves to correct for recognition errors is likely a better path forward.  

Recent research has shown that CNNs are able to predict neural responses in the visual cortex of primates~\cite{yamins2014performance}. This, coupled with excellent benchmark dataset results across multiple recognition domains, suggests that good progress is being made towards reaching human-like performance. As a strong counterpoint, our psychophysics experiments show that the current most popular CNN models sometimes fail to correctly classify images that humans do not make mistakes on. What is missing from the models that is causing this behavioral discrepancy? With psychophysics as a guide, we can more easily discover what is missing --- making it harder for us to be fooled by the person inside of the Chinese room.


%

\ifCLASSOPTIONcompsoc
  \section*{Acknowledgments}
\else
  \section*{Acknowledgment}
\fi

The authors thank Lucas Parzianello for helping import Blendswap models into PsyPhy and Brian Turnquist for providing feedback on an early draft of this work. Funding was provided under IARPA contract \#D16PC00002, NSF DGE \#1313583, and NSF SBIR Award \#IIP-1621689. Hardware support was generously provided by the NVIDIA Corporation.

\ifCLASSOPTIONcaptionsoff
  \newpage
\fi



%



{\small
\bibliographystyle{IEEEtran}
\bibliography{egbib}
}

%








\end{document}